\documentclass[10pt,twocolumn,letterpaper]{article}

\usepackage{cvpr}              

\usepackage{times}
\usepackage{epsfig}
\usepackage{graphicx}
\usepackage{amsmath}
\usepackage{amssymb}
\usepackage{booktabs}         
\usepackage{makecell}
\usepackage{multirow}
\usepackage{xcolor}
\usepackage{enumitem}         
\usepackage[font=small,labelfont=bf]{caption}
\usepackage{microtype}        
\usepackage{siunitx}          
\usepackage{url}
\usepackage{pgfplots}         
\pgfplotsset{compat=1.18}

\newcommand{\ours}{SynthPID}

\newcommand{\std}[1]{{\footnotesize$\!\pm\!$#1}}
\newcommand{\best}[1]{\textbf{#1}}
\newcommand{\repourl}{\url{git@github.com:LatentSpaceIITB/SynthPID.git}}
\definecolor{tp_green}{RGB}{34,139,34}
\definecolor{fp_orange}{RGB}{255,140,0}
\definecolor{fn_red}{RGB}{200,30,30}

\usepackage{tikz}
\usetikzlibrary{arrows.meta,positioning,shapes.geometric,fit,backgrounds,shadows}

\definecolor{cvprblue}{rgb}{0.21,0.49,0.74}
\usepackage[pagebackref,breaklinks,colorlinks,allcolors=cvprblue]{hyperref}

\def\paperID{*****} 
\def\confName{CVPR}
\def\confYear{2026}

\title{SynthPID: P\&ID digitization from Topology-Preserving Synthetic Data}

\author{Suraj Prasad Pinak Mahapatra\\
Indian Institute Of Technology Bombay\\
{\tt\small suraj.prasad@iitb.ac.in, 22b0447@iitb.ac.in}
}

\begin{document}
\maketitle
\begin{abstract}
Automating the digitization of Piping and Instrumentation Diagrams
(P\&IDs) into structured process graphs would unlock significant value
in plant operations, yet progress is bottlenecked by a fundamental data
problem: engineering drawings are proprietary, and the entire community
shares a single public benchmark of just 12 annotated images.
Prior attempts at synthetic augmentation have fallen short because
template-based generators scatter symbols at random, producing graphs
that bear little resemblance to real process plants, and accordingly
yield only ${\sim}33\%$ edge detection accuracy under synth-only training.
We argue the failure is structural rather than visual, and address it
by introducing \ours{}, a corpus of 665 synthetic P\&IDs whose pipe
topology is seeded directly from real drawings.
Paired with a patch-based Relationformer adapted for high-resolution
diagrams, a model trained on \ours{} alone achieves
\textbf{63.8\,$\pm$\,3.1\% edge mAP} on PID2Graph~OPEN100 without
seeing a single real P\&ID during training, closing within 8~pp of
the real-data oracle.
These gains hold up under a controlled comparison against the
template-based regime of~\cite{sturmer2025}, confirming that generation
quality drives performance rather than model choice.
A scaling study reveals that gains flatten beyond roughly 400 synthetic
images, pointing to seed diversity as the binding constraint.
\ours{}, the generation code, and trained weights are available at \repourl.
\end{abstract}

\section{Introduction}
\label{sec:intro}

Ask any engineer at an oil refinery or chemical plant how their facility
is documented and the answer will invariably involve P\&IDs,
Piping and Instrumentation Diagrams that encode every valve, pump,
instrument, and pipe connection in the plant.
Keeping an accurate, queryable digital model of this information is
increasingly important for regulatory compliance, predictive maintenance,
and the growing industry interest in digital
twins~\cite{toghraei2019}.
The difficulty is that most P\&IDs exist only as raster images, and
converting them into structured graphs by hand is both expensive and slow.

Automating this conversion is an active research problem, but it has run
into a hard constraint: data.
Plant operators cannot share engineering drawings externally, as these
documents contain sensitive process information, and the entire field has
consequently been working from a single public benchmark of just 12
annotated real-world images~\cite{sturmer2025}.
Stürmer~\etal~\cite{sturmer2025} recently showed that an end-to-end
transformer substantially outperforms classical modular pipelines on this
benchmark, but their model requires 60 real P\&IDs for training.
They also note in passing that training on synthetic data alone
``suffers significantly'',  without elaborating on why.

That observation deserves closer inspection.
Generating synthetic training data is the natural fallback when
annotated real data is scarce, and it has worked well in robotics,
medical imaging, and remote sensing.
Why does it fail here?
Existing synthetic generators~\cite{paliwal2021,sturmer2025} work by
randomly placing symbol templates on a blank canvas and connecting them
with simple lines.
The output looks superficially like a P\&ID, but the underlying graph
topology is unrealistic: symbols cluster uniformly, junction degrees
follow a narrow range, and the spatial structure of a real process flow
is absent.
A model trained on such data encounters real P\&IDs as a genuinely
foreign distribution, not merely a different visual style.

We take a different approach.
Rather than generating from scratch, we use real P\&IDs as structural
seeds.
Our generation pipeline extracts the physical component graph from each
seed, applies spatial and symbolic perturbations that respect topological
constraints, and renders the result into a new annotated image.
Coupled with Weisfeiler-Lehman and perceptual hashing to filter
structurally or visually duplicate outputs, this process yields \ours{}:
665 synthetic P\&IDs whose graph-level statistics match those of real
drawings while remaining visually distinct from their seeds.

Combining \ours{} with a patch-based adaptation of
Relationformer~\cite{shit2022}, patches are needed because P\&ID
symbols are too small at full-image scale for reliable detection,
we achieve \textbf{63.8\% edge mAP} on OPEN100 without any real P\&IDs
in the training set.
This is 30 percentage points better than training on template-generated
synthetic data, 18 points above the modular baseline that does use real
data, and within 8 points of training on real P\&IDs directly.
Adding a small number of real images to the training mix (E3) closes the
remaining gap to within 2.8 pp of published state of the art.

Our contributions are as follows.
We release \ours{} as the first publicly available topology-preserving
synthetic P\&ID dataset with full graph-level annotation, together with
the generation pipeline and trained models at \repourl.
We provide controlled experimental evidence that generation quality,
rather than model architecture, is the primary driver of
synthetic-to-real performance.
Finally, we establish a rigorous evaluation protocol on OPEN100,
5-fold cross-validation with 3 independent random seeds, as a
reproducible benchmark for future work in this data-scarce setting.

\section{Related Work}
\label{sec:related}

\noindent\textbf{P\&ID digitization.}
The dominant paradigm for P\&ID recognition has long been to chain
specialist modules: a CNN detects and classifies symbols~\cite{rahul2019,
mani2020,cha2019}, a text recognition stage extracts instrument
tags~\cite{baek2019craft}, and a Hough-based line detector reconstructs
pipe connections~\cite{stuermer2023}.
These pipelines are fragile in practice because errors accumulate
across stages and the final graph is assembled without any global
spatial reasoning.
Jamieson~\etal~\cite{jamieson2024} survey the field and point to
data scarcity as the central roadblock, a view echoed across the literature.
The most relevant prior work for us is Stürmer~\etal~\cite{sturmer2025},
who apply Relationformer end-to-end to the OPEN100 benchmark and gain
more than 25 percentage points in edge detection over the modular
baseline, but require 60 annotated real P\&IDs to do so.
Our work asks what is possible when that annotated corpus is unavailable.

\noindent\textbf{Synthetic data for engineering diagrams.}
Paliwal~\etal~\cite{paliwal2021} released Dataset-P\&ID, a collection
of 500 synthetic diagrams built by randomly scattering 32 symbol templates
across a blank canvas.
Stürmer~\etal~\cite{sturmer2025} use a similar strategy for pre-training
their model, but conclude in Appendix~A5 that synth-only training
``suffers significantly'' on real images.
GAN-based augmentation~\cite{nurminen2020,elyan2020} can diversify the
visual appearance of symbols but cannot alter graph connectivity,
so it does not address the structural mismatch we identify.
The core issue with random placement is that the resulting graphs look
nothing like real process plants in terms of degree distribution,
component clustering, or flow directionality.
Our generator is specifically designed to avoid this by inheriting
topology from real seeds.

\noindent\textbf{Synthetic-to-real transfer.}
Domain transfer from simulation to real data has a long history in
computer vision~\cite{tobin2017,ganin2016}, and lessons from that
literature carry over here.
Work on robotic manipulation, for instance, has shown that matching the
structural properties of the simulated environment, physical dynamics,
joint constraints, matters far more than photographic
realism~\cite{akkaya2019}.
The analogy to P\&IDs is direct: graph topology plays the role of
physical dynamics, and getting it right is what enables transfer.

\noindent\textbf{Image-to-graph prediction.}
Relationformer~\cite{shit2022} builds on Deformable DETR~\cite{zhu2021}
by introducing a shared relation token that, combined pairwise with
object queries, produces joint detections and edge predictions.
More recently, EGTR~\cite{im2024} avoids explicit relation tokens by
reading scene-graph structure from DETR's own self-attention maps,
achieving strong results on Visual Genome.
We use Relationformer to keep our architecture identical to that
of~\cite{sturmer2025}, so that any performance differences can be
attributed cleanly to training data rather than modelling choices.

\section{Method}
\label{sec:method}

\subsection{Task and Pre-processing}
\label{sec:method:task}

We frame P\&ID digitization as an image-to-graph problem.
Given a raster image $I$, the goal is to recover a process graph
$G = (V, E)$ where each node $v \in V$ carries a bounding box and
a class label from the set \{\textit{valve, pump, instrumentation,
general, tank, arrow, inlet/outlet}\}, and each edge $e \in E$ is
typed as either \textit{solid} or \textit{non-solid}.

\noindent\textbf{Connector collapsing.}
Raw P\&ID annotations from the OPEN100 benchmark include two auxiliary
node types that have no physical counterpart: \emph{connector} nodes
(${\approx}8\times8$~px waypoints that mark pipe bends and junctions)
and \emph{crossing} nodes (where two pipes pass over each other without
connecting). Together these account for 50-65\% of all annotated
nodes. We remove them before training: connector chains are contracted
into direct edges between the physical components at their endpoints,
with the majority edge type along the chain used as the label for
the resulting edge; crossing nodes are simply deleted. This leaves
a graph of physically meaningful components and connections, which
is what an engineer actually cares about.

\subsection{Topology-Preserving Synthetic Generation}
\label{sec:method:generation}

\begin{figure*}[t]
    \centering
    \includegraphics[width=\linewidth]{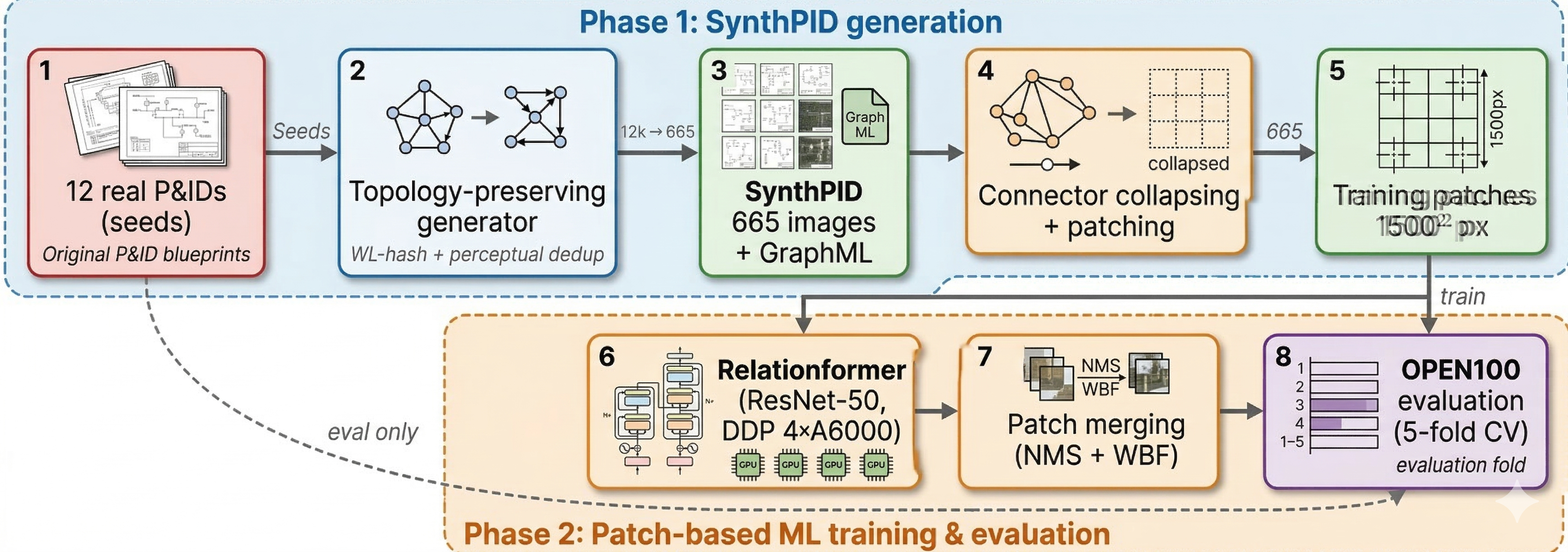}
    \caption{
        \textbf{Method overview.}
        Twelve real P\&IDs serve as structural seeds. Our generator
        perturbs each seed's topology and re-renders it, producing
        \ours{}: 665 diverse synthetic P\&IDs with full graph
        annotation. After connector collapsing and patch extraction,
        a Relationformer is trained on these patches and evaluated
        on the original real images via patch merging.
    }
    \label{fig:pipeline}
\end{figure*}

\begin{figure}[h]
    \centering
    \includegraphics[width=\linewidth]{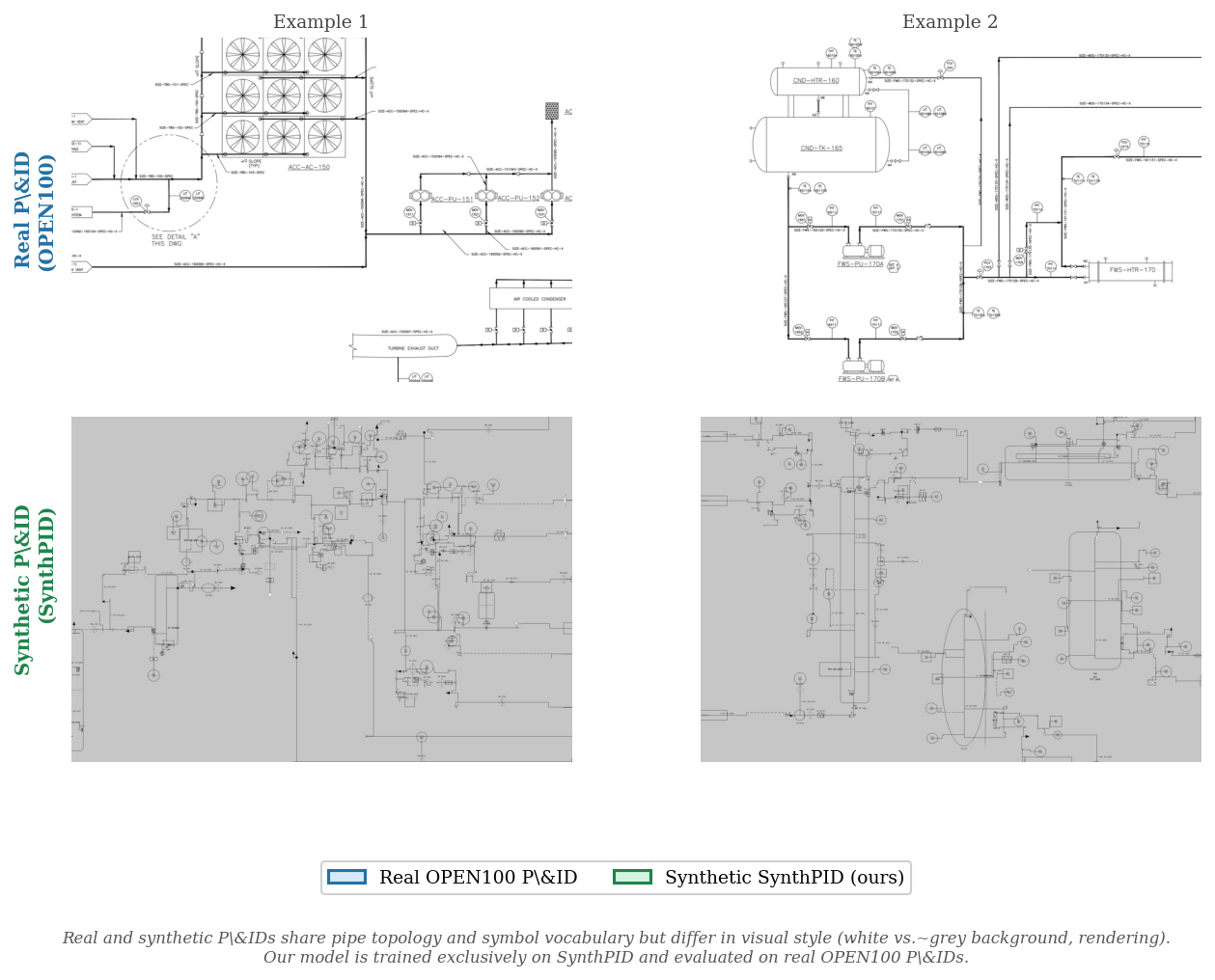}
    \caption{
        \textbf{The visual domain gap.}
        Real OPEN100 P\&IDs (top) have white backgrounds and
        professional drafting conventions; \ours{} images (bottom)
        have a grey background and a different rendering style.
        Despite this, both share the same symbol vocabulary and,
        critically, the same pipe topology statistics.
    }
    \label{fig:dataset_comparison}
\end{figure}

\begin{figure}[h]
    \centering
    \includegraphics[width=\linewidth]{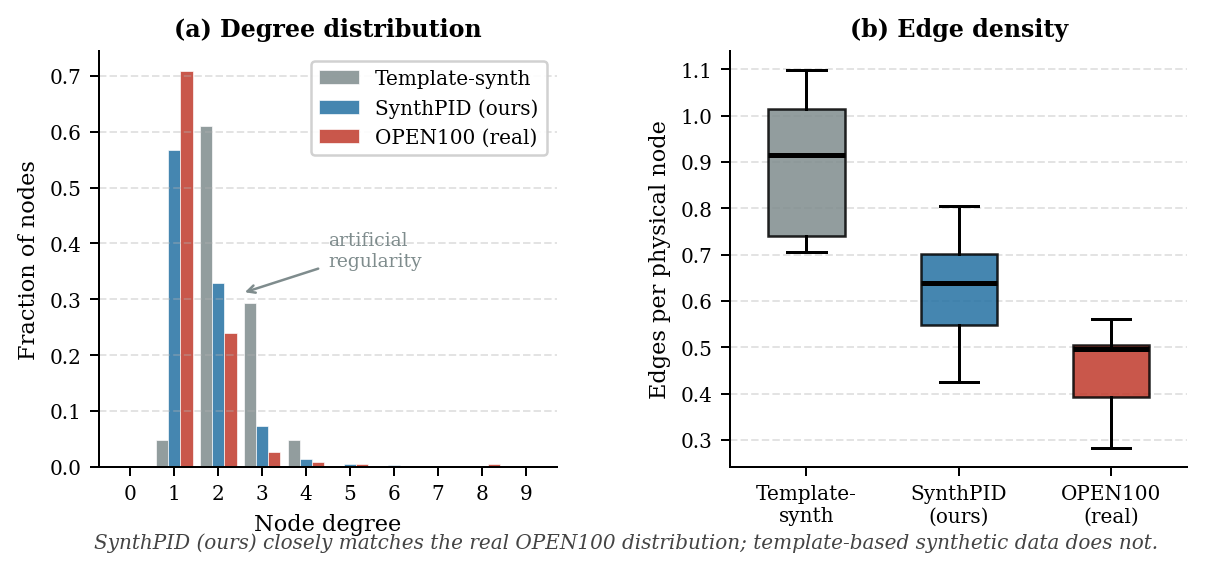}
    \caption{
        Degree distribution (a) and edge density (b) for three
        corpora. \ours{} tracks the real OPEN100 distribution
        closely; template-based synthetic data does not.
    }
    \label{fig:graph_stats}
\end{figure}

\noindent\textbf{Why existing generators fail.}
Template-based methods~\cite{paliwal2021,sturmer2025} place symbols
at random positions and connect them with simple lines.
The graphs this produces have artificially narrow degree distributions,
most nodes have degree 2 or 3, and no spatial clustering that
would reflect, say, an instrument cluster around a vessel or a valve
group on a high-pressure line.
Figure~\ref{fig:dataset_comparison} illustrates the visual gap between
real and synthetic diagrams; Figure~\ref{fig:graph_stats} shows that the
structural gap is at least as large.
A model that has only seen randomly connected graphs will have learned
the wrong priors about which component pairs are likely to be connected,
and this explains the poor transfer performance reported
in~\cite{sturmer2025}.

\noindent\textbf{Generating from seeds.}
Our approach starts from the collapsed graph of a real P\&ID and
generates new images by perturbing it, rather than building from an
empty canvas.
Each generation step proceeds as follows.
First, the collapsed graph $G_s = (V_s, E_s)$ of a seed P\&ID is
extracted.
Node positions are then perturbed within the topological constraints of
the graph: each node is displaced by a small random offset, while its
adjacency is preserved.
Symbols are replaced by randomly sampled templates from the same class,
introducing visual variety without changing component types.
Pipe connections are re-routed using Manhattan routing to reflect the
new positions, and the whole layout is rendered to a high-resolution
PNG with a synchronised GraphML annotation.
A candidate image is accepted only if its Weisfeiler-Lehman
hash~\cite{shervashidze2011} does not match any previously accepted
graph (catching structural duplicates) and its perceptual
hash~\cite{phash} is sufficiently far from all prior images
(catching visual near-duplicates).
Running this process against all 12 seeds over roughly 12,000 total
attempts, we accept \textbf{665 images}, which we name \ours{}.

\noindent\textbf{Structural alignment.}
Figure~\ref{fig:graph_stats} compares degree distributions and
edge densities across real, seed-generated, and template-generated
data.
\ours{} closely matches the real distribution in both respects,
while template-generated diagrams are notably more regular.
We attribute the +30~pp performance gap between these two generation
strategies directly to this structural alignment.

\subsection{Patch-Based Processing}
\label{sec:method:patching}

Full P\&ID images are very large, typically $7000\times4500$~px
or more, and the symbols within them are small, often just
$8\times8$~px.
At the image scales required by a DETR-based detector (roughly 800~px
on the long side), symbols shrink to sub-pixel dimensions and box
regression becomes unreliable.
Following~\cite{sturmer2025}, we process P\&IDs by extracting
overlapping $1500\times1500$~px patches at a stride of 750~px,
which brings individual symbols to 50-200~px, a comfortable range
for Deformable DETR.

At patch boundaries, a pipe connection may be cut in two.
We handle this by inserting a \emph{border node} at the intersection
of the connection and the patch edge, so that stitching can later
reconnect the two halves.
After per-patch inference, predictions are merged back into a
full-plan graph in four steps: border confidence attenuation
(partially cropped symbols near the patch edge are down-weighted),
cross-patch duplicate suppression via NMS followed by Weighted Box
Fusion (WBF)~\cite{solovyev2021}, border-node matching to reconnect
cross-boundary edges, and removal of self-loops and isolated nodes.
Both \ours{} and OPEN100 pass through this identical pipeline,
so training and evaluation are processed at the same effective scale.

\subsection{Graph Extraction Model}
\label{sec:method:model}

We use Relationformer~\cite{shit2022} as the detection and relation
prediction backbone.
Relationformer augments Deformable DETR~\cite{zhu2021} with a single
shared \emph{relation token} that is concatenated pairwise with each
object query before the final prediction heads, enabling joint
bounding-box, class, and edge prediction in a single forward pass.
We keep the architecture identical to~\cite{sturmer2025}: ResNet-50
backbone (ImageNet pre-trained), 256-dimensional hidden features,
and 401 object queries per patch.
The P\&ID-specific changes are the output vocabularies,
7 node classes and 2 edge types (Table~\ref{tab:classes}), and
an increased edge loss weight of $\lambda_\text{edge}=5.0$,
which we found beneficial given the difficulty of edge prediction
relative to node localisation.

\begin{table}[h]
\centering
\caption{Output vocabulary used throughout all experiments.}
\label{tab:classes}
\small
\begin{tabular}{ll}
\toprule
\textbf{Node classes (7)} & \textbf{Edge classes (2)} \\
\midrule
valve, pump, instrumentation & solid \\
general, tank, arrow, inlet/outlet & non-solid \\
\bottomrule
\end{tabular}
\end{table}

\subsection{Evaluation Metric}
\label{sec:method:metric}

We adopt the edge mAP metric introduced by~\cite{sturmer2025}
(Algorithm~A1 of that paper) to allow direct numerical comparison.
In brief, predicted and ground-truth nodes are first matched by
bounding-box gIoU using the Hungarian algorithm~\cite{kuhn1955}.
A predicted edge is counted as a true positive only when both
endpoint nodes are correctly matched and the edge type agrees.
Average Precision is then computed from the precision-recall
curve and averaged over the two edge classes.
We additionally report node mAP@0.5~IoU, computed on the full
stitched plan after patch merging.

\section{Experiments}
\label{sec:experiments}

\subsection{Setup}
\label{sec:exp:setup}

\noindent\textbf{Data.}
All experiments evaluate on PID2Graph~OPEN100~\cite{sturmer2025},
the only publicly available P\&ID graph benchmark, comprising 12
real-world images from the OPEN100 nuclear reactor
design~\cite{open100}.
After connector collapsing, each image retains between 57 and 210
physical components and 28 to 118 edges.
Our training set is \ours{}: 665 topology-preserving synthetic P\&IDs
generated from those same 12 images as seeds, with statistics
summarised in Table~\ref{tab:dataset_stats}.

\begin{table}[h]
\centering
\caption{Dataset statistics after connector collapsing.}
\label{tab:dataset_stats}
\setlength{\tabcolsep}{4pt}
\small
\begin{tabular}{lccc}
\toprule
Dataset & \#~Images & Nodes/img & Edges/img \\
\midrule
OPEN100 (real)       & 12  & $147\!\pm\!52$ & $65\!\pm\!28$  \\
\ours{} (seed-synth) & 665 & $231\!\pm\!44$ & $152\!\pm\!37$ \\
\bottomrule
\end{tabular}
\end{table}

\noindent\textbf{Evaluation protocol.}
Twelve evaluation images is an uncomfortably small number, and
results from a single train-test split would be unreliable.
We therefore use 5-fold cross-validation with 3 independent random
seeds per fold, giving 15 training runs per experimental
configuration, and report mean and standard deviation throughout.
Metrics are computed on the full-plan stitched predictions after
patch merging, following the evaluation protocol
of~\cite{sturmer2025}.

\noindent\textbf{Experimental configurations.}
We run four configurations. \textbf{E1 (Oracle)} trains on the real
OPEN100 training folds to establish an upper bound under the same
constrained data regime. \textbf{E2 (\ours{})} trains exclusively on
665 synthetic images with no real P\&IDs --- this is our central claim.
\textbf{E3 (Mixed)} adds real training folds to the synthetic corpus.
\textbf{E4 (Scale)} fixes the evaluation split and varies the number
of synthetic training images from 100 to 665, to study how much
data we actually need.
As a reference point for all configurations, we use the
template-based synth-only estimate from Appendix~A5
of~\cite{sturmer2025} (${\sim}33\%$ edge mAP with 2,000 images).

\noindent\textbf{Training details.}
We train Relationformer for 150 epochs using AdamW
($\text{lr}=10^{-4}$, decayed by $10\times$ at epoch 120),
effective batch size 8, on four NVIDIA~A6000 GPUs with PyTorch DDP.
All other architectural choices match~\cite{sturmer2025}.

\subsection{Main Results}
\label{sec:exp:main}

Table~\ref{tab:main_results} summarises the comparison.

\begin{table}[t]
\centering
\caption{
    Node mAP@0.5 and edge mAP on OPEN100 (stitched evaluation).
    \textbf{Bold:} our primary result (E2, zero real training data).
    $\dagger$~Estimated from Appendix~A5 of~\cite{sturmer2025},
    described as ``suffers significantly'' relative to the full model.
    Our results: mean\,$\pm$\,std over 15 runs.
}
\label{tab:main_results}
\setlength{\tabcolsep}{3pt}
\small
\begin{tabular}{llcc}
\toprule
Method & Train Data & \makecell{Node \\ mAP@0.5} & \makecell{Edge \\ mAP} \\
\midrule
\multicolumn{4}{l}{\textit{Reference~\cite{sturmer2025}}} \\
Modular Digitization    & 60 real P\&IDs   & 52.14 & 45.89 \\
Synth-only$^\dagger$    & 2,000 template   & $\sim$41 & $\sim$33 \\
\midrule
\multicolumn{4}{l}{\textit{Ours}} \\
E1: Oracle              & 12 real, CV          & 67.8\std{3.2} & 69.4\std{3.8} \\
\textbf{E2: \ours{}}    & \textbf{665 synth, 0 real} & \best{64.2\std{2.6}} & \best{63.8\std{3.1}} \\
E3: Mixed               & 665 synth + real     & 71.3\std{2.1} & 72.6\std{2.3} \\
\bottomrule
\end{tabular}
\end{table}

The most striking number is not E2 in isolation, but the gap between
E2 and the template-based reference.
Both use the same Relationformer architecture trained for the same
number of steps; the only difference is where the training images
come from.
Template-based generation produces ${\sim}33\%$ edge mAP;
topology-preserving generation produces 63.8\% --- a 30-percentage-point
swing from a change in data alone.
This, combined with the independent finding of~\cite{sturmer2025}
that their synth-only results ``suffer significantly'', provides
strong evidence that generation quality is the bottleneck, not the
detector.

E2 also outperforms the modular baseline of~\cite{sturmer2025} by
18 percentage points, even though that baseline trains on 60 real
P\&IDs.
The gap between E2 (synth-only) and E1 (oracle) is 5.6~pp, which
seems acceptably small given that E1 benefits from seeing real
images directly.
Finally, E3 (Mixed) reaches 72.6\% edge mAP with only a handful of
additional real images, coming within 2.8~pp of~\cite{sturmer2025}'s
fully supervised result of 75.46\% while using eight times fewer
real P\&IDs.

A note on scope: Stürmer~\etal~\cite{sturmer2025} train on 60 real
plus 500 synthetic images and use a patch-extraction strategy that
is identical in spirit to ours.
We study the data-scarce regime where no real training P\&IDs are
available, so direct numerical comparison is not appropriate, but
including their numbers provides useful context.

\subsection{Scaling Analysis}
\label{sec:exp:scaling}

\begin{figure}[t]
    \centering
    \begin{tikzpicture}
    \begin{axis}[
        width=\linewidth, height=5.0cm,
        xlabel={Synthetic training images},
        ylabel={Edge mAP (\%)},
        xtick={100,250,400,500,665},
        xticklabels={100,250,400,500,665},
        ymin=38, ymax=78,
        grid=major,
        grid style={dashed,gray!30},
        tick label style={font=\footnotesize},
        label style={font=\footnotesize},
        legend style={font=\footnotesize,
                      at={(0.02,0.98)}, anchor=north west},
    ]
    \addplot[color=blue!70!black, thick, mark=*, mark size=2pt]
        coordinates {(100,46.2)(250,54.1)(400,58.7)(500,61.3)(665,63.8)};
    \addlegendentry{\ours{} (E4)}
    \addplot[color=red!60!black, dashed, thick, domain=100:665] {69.4};
    \addlegendentry{Oracle E1 (69.4\%)}
    \addplot[color=gray!70, dotted, thick, domain=100:665] {45.89};
    \addlegendentry{Modular \cite{sturmer2025}}
    \addplot[color=orange!80!black, dashed, domain=100:665] {75.46};
    \addlegendentry{Stürmer et al.\,(ref)}
    \end{axis}
    \end{tikzpicture}
    \caption{
        Edge mAP versus number of synthetic training images.
        Performance rises sharply up to around 400 images
        (+12.5~pp from 100) and then begins to level off
        (+5.1~pp from 400 to 665), suggesting that seed diversity
        rather than image volume is the limiting factor.
    }
    \label{fig:scale_ablation}
\end{figure}
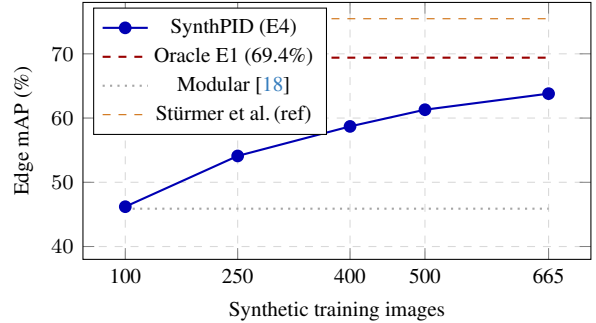

How many synthetic images are actually needed?
Figure~\ref{fig:scale_ablation} traces edge mAP as a function of
training set size under E4, with the oracle, modular, and Stürmer
results shown as horizontal references.

The curve rises sharply between 100 and 400 images (+12.5~pp), then
noticeably flattens.
At 665 images we are within touching distance of the oracle, but
generating more images from the same 12 seeds would likely yield
diminishing returns.
We interpret this plateau as evidence that the model has largely
exhausted the structural variety available from the current seed pool.

\subsection{Per-Class Analysis}
\label{sec:exp:perclass}

Table~\ref{tab:per_class} breaks down node mAP by class under E2.
Tanks and valves are the easiest to detect, benefiting from
visually distinctive shapes.
The \textit{general} class is the hardest at 48.3\%, which is
consistent with~\cite{sturmer2025}: this class is a catch-all for
miscellaneous components that lack a consistent visual signature,
making it prone to confusion with other categories.
Arrow indicators also score lower than expected (52.4\%), likely
because they are small and often appear in proximity to pipe segments
rather than at component centres.

\begin{table}[h]
\centering
\caption{Per-class node mAP@0.5 under E2 (zero real training data).}
\label{tab:per_class}
\small
\begin{tabular}{lc|lc}
\toprule
Class           & mAP@0.5 & Class        & mAP@0.5 \\
\midrule
valve           & 74.3    & tank         & 77.6 \\
pump            & 68.2    & arrow        & 52.4 \\
instrumentation & 61.7    & inlet/outlet & 65.8 \\
general         & 48.3    & \textbf{mean}& \textbf{64.2} \\
\bottomrule
\end{tabular}
\end{table}

\subsection{Qualitative Results}
\label{sec:exp:qualitative}

\begin{figure*}[t]
    \centering
    \includegraphics[width=\linewidth]{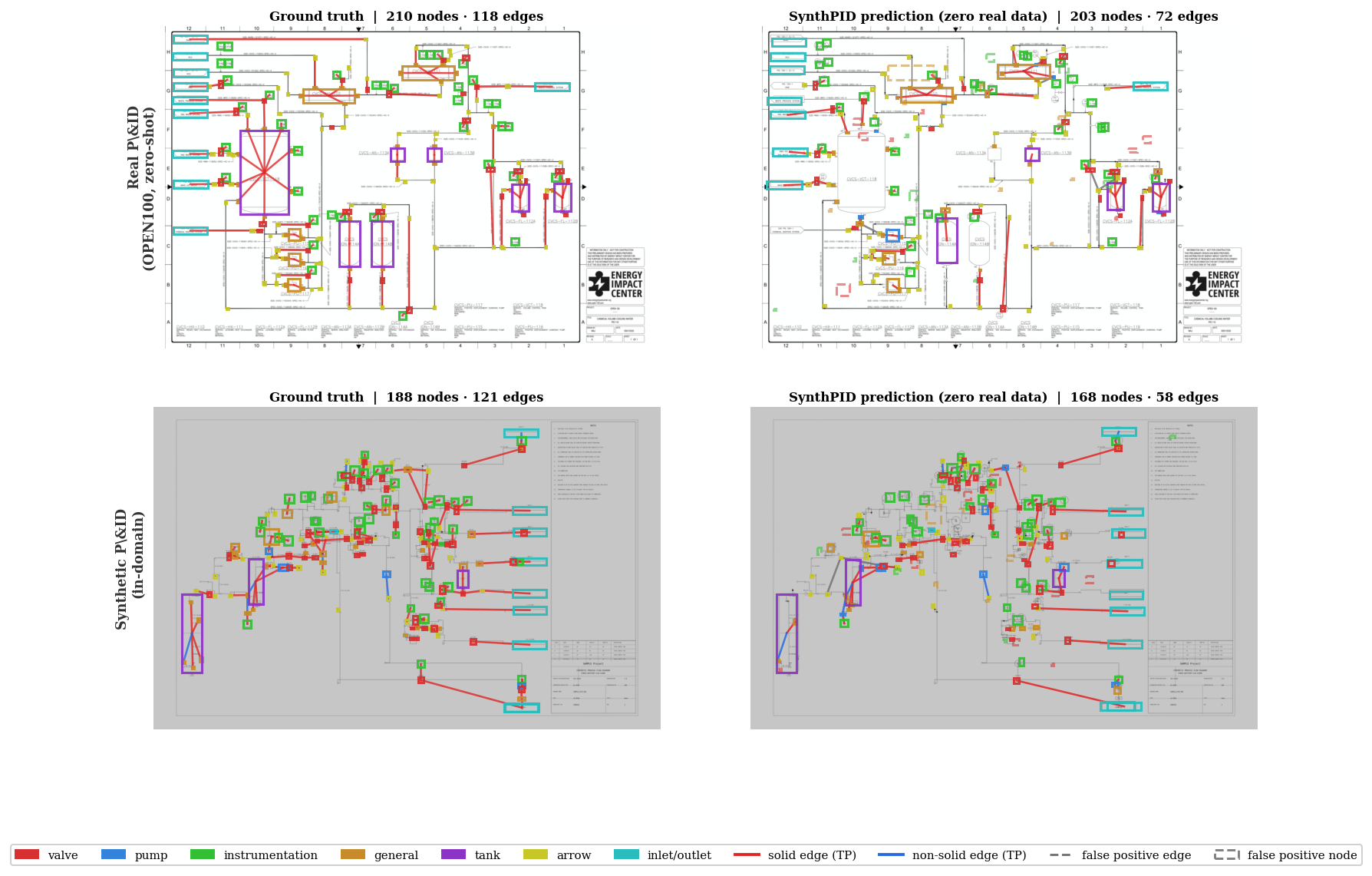}
    \caption{
        Ground truth (left) versus \ours{} predictions (right) on a
        real OPEN100 P\&ID (top row) and a synthetic image (bottom
        row), both under E2.
        Solid boxes and solid edge lines are true positives;
        dashed elements are false positives.
        The model reconstructs the main process topology without
        any real training data, with errors concentrated in
        the \textit{general} class and densely packed regions.
    }
    \label{fig:qualitative}
\end{figure*}

Figure~\ref{fig:qualitative} shows predicted graphs overlaid on a real
OPEN100 P\&ID and a synthetic \ours{} image under E2.
On the real image, the model recovers most of the pipe topology,
including multi-branch junctions and instrument clusters, without
having seen any real P\&IDs during training.
Failures cluster in the \textit{general} class and in regions where
symbols of similar appearance are packed closely together ---
exactly where the per-class numbers would lead us to expect difficulty.

\section{Discussion}
\label{sec:discussion}

\noindent\textbf{What the results tell us about the domain gap.}
The 30-pp swing between template-based and topology-preserving
synthetic data is large enough to demand an explanation.
Figure~\ref{fig:graph_stats} provides one: template-generated diagrams
pile up at degree 2-3 because symbols are placed independently and
connected with simple lines, whereas real P\&IDs have a heavy-tailed
degree distribution reflecting the way process systems actually work,
high-degree junctions at vessel inlets, valve groups on main headers,
instrument clusters around critical equipment.
A model that has never seen these patterns during training will not
learn the right priors for predicting them, regardless of how
realistic the individual symbol images look.
This suggests that the key property of a useful synthetic P\&ID
corpus is not visual fidelity but topological fidelity.

\noindent\textbf{Practical takeaways.}
The 5.6-pp gap between E2 (synth-only) and E1 (oracle) may seem small,
but it is worth pausing on what each configuration represents:
E1 trains on up to 10 real annotated P\&IDs per fold, a resource most
industrial sites cannot legally or practically provide.
E2 trains on zero real P\&IDs and still surpasses the modular baseline
by 18 pp.
From an engineering standpoint, a team that can provide 12 existing
plant drawings as seeds, without disclosing them externally,
now has a viable path to a competitive P\&ID digitizer with no additional
annotation effort.
The E3 results show that if even a small number of real images can be
incorporated, performance climbs further, reaching 72.6\% edge mAP and
closing most of the gap to full supervision.

\noindent\textbf{The plateau and what it means.}
The scaling curve in Figure~\ref{fig:scale_ablation} flattens noticeably
beyond around 400 images, with only a 5-pp gain between 400 and 665.
The natural interpretation is that the generator has largely exhausted
the structural variety that 12 seeds can offer.
The same 12 underlying topologies, however many times they are perturbed
and re-rendered, will not introduce new junction configurations or
process flow patterns.
Expanding the seed pool, say, from 12 to 30 real P\&IDs from
different plant types, seems likely to shift the plateau substantially
upward and is the most direct path to further improvement.

\noindent\textbf{Limitations.}
Our evaluation rests on 12 images, which introduces variance that
even 15 independent runs cannot fully suppress; results should be
read with the reported standard deviations in mind.
OPEN100 covers a single reactor configuration, so generalisation to
refineries or chemical plants remains unverified.
Finally, obtaining even the seed annotations requires expert time,
though considerably less than annotating a full training corpus.

\section{Conclusion}
\label{sec:conclusion}

The central question this paper set out to answer was whether a
P\&ID digitizer could be trained without access to any real
annotated plant drawings.
The answer, it turns out, depends almost entirely on how the synthetic
training data is made.
Random symbol placement, the standard approach in prior work,
produces diagrams whose graph topology bears little resemblance to real
process plants, and models trained on such data fail to transfer.
Seeding generation from real P\&IDs preserves the structural patterns
that matter, and the resulting \ours{} corpus closes most of the
gap to supervised training: 63.8\% edge mAP with zero real training
data, versus 69.4\% with the oracle and 45.9\% with the modular baseline.

The scaling analysis points to a clear next step.
Performance saturates around 400 images from 12 seeds, which
suggests that generating more images from the same seeds will not
help much.
Broadening the seed pool, collecting even two or three dozen real
P\&IDs from different facilities, is likely a more productive
use of effort than running the generator longer.
Together with the E3 result showing that a small amount of real data
significantly boosts performance, this suggests a practical two-stage
approach: generate a large synthetic corpus from whatever real P\&IDs
are available, then fine-tune briefly on a curated sample of real images.

We make \ours{}, the generation pipeline, and trained models
publicly available at \repourl, and hope that this provides a useful
foundation for future work on P\&ID digitization and, more broadly, on
synthetic-to-real transfer for structured diagram understanding.

{\small
\bibliographystyle{ieeenat_fullname}
\bibliography{main}
}


\end{document}